\documentclass[10pt,twocolumn,letterpaper]{article}
\usepackage{caption}
\usepackage{amssymb}
\usepackage{cvpr}
\usepackage{times}
\usepackage{epsfig}
\usepackage{graphicx}
\usepackage{amsmath}
\usepackage{amssymb}
\usepackage{booktabs}
\usepackage{color, colortbl}
\usepackage[numbers,sort,compress]{natbib} % sort citations
\definecolor{salmon}{rgb}{1.0, 0.90, 0.75}
\definecolor{LightCyan}{rgb}{0.88,1,1}
\definecolor{azure}{rgb}{0.94, 1.0, 1.0}
\definecolor{khaki}{rgb}{0.94, 0.9, 0.55}
\definecolor{lemonchiffon}{rgb}{1.0, 0.98, 0.8}
\definecolor{mistyrose}{rgb}{1.0, 0.89, 0.88}
\definecolor{palerobineggblue}{rgb}{0.59, 0.87, 0.82}
\definecolor{mossgreen}{rgb}{0.68, 0.87, 0.68}
\definecolor{magicmint}{rgb}{0.77, 1, 0.92}
\definecolor{grannysmithapple}{rgb}{0.66, 0.89, 0.63}	\definecolor{lavenderblue}{rgb}{0.9, 0.9, 1.0}
\definecolor{peachorange}{rgb}{1.0, 0.8, 0.6}
\definecolor{pistachio}{rgb}{0.85, 1, 0.72}

\usepackage[pagebackref=true,breaklinks=true,letterpaper=true,colorlinks,bookmarks=false]{hyperref}
\cvprfinalcopy % *** Uncomment this line for the final submission

\def\cvprPaperID{1426} % *** Enter the CVPR Paper ID here

\ifcvprfinal\fi

\title{\textit{Speech2Action:} Cross-modal Supervision for Action Recognition}

\author{Arsha Nagrani$^{1,2}$~~~~Chen Sun$^2$~~~~David Ross$^2$\\ ~~~~Rahul Sukthankar$^2$~~~~Cordelia Schmid$^{2}$ ~~~~Andrew Zisserman$^{1,3}$
\\
\small{$^1$VGG, Oxford~~~~$^2$Google Research~~~~$^3$DeepMind~~~~}\\
\small{\url{https://www.robots.ox.ac.uk/~vgg/research/speech2action/}}
}

\begin{document}
\twocolumn[\maketitle\vspace{-3em}
\begin{center}
  \includegraphics[width=1\linewidth]{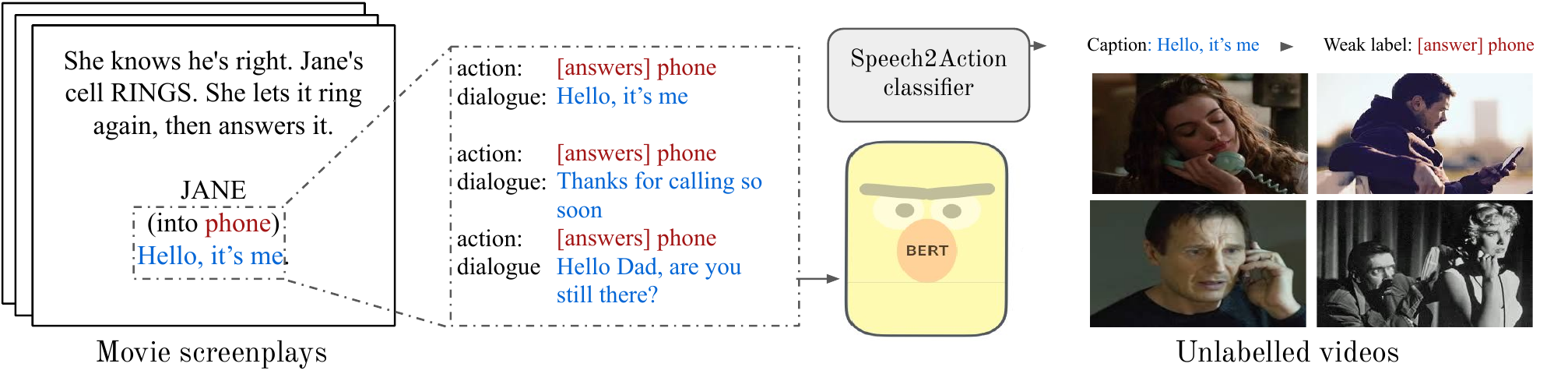}
\end{center}
\captionof{figure}{\textbf{Weakly Supervised Learning of Actions from Speech Alone:} The co-occurrence of speech and scene descriptions in movie screenplays (text) is used to learn a \texttt{Speech2Action} model that predicts actions from transcribed speech \textit{alone}.  Weak labels for visual actions can then be obtained by applying this model to the speech in a large~\textit{unlabelled} set of movies. }\label{fig:teaser}\medbreak]
%\thispagestyle{empty}
% \maketitle
%%%%%%%%% ABSTRACT
\begin{abstract}
Is it possible to guess human action from dialogue alone? 
In this work we investigate the link between spoken words and actions in movies. We note that movie screenplays describe actions, as well as contain the speech of characters and hence can be used to learn this correlation with no additional supervision. We train a BERT-based \texttt{Speech2Action} classifier on over a thousand movie screenplays, to predict action labels from transcribed speech segments. 

We then apply this model to the speech segments of a large \texttt{unlabelled} movie corpus (188M speech segments from 288K movies). Using the predictions of this model, we obtain weak action labels for over 800K video clips. By training on these video clips, we demonstrate superior action recognition performance on standard action recognition benchmarks, without using a single manually labelled action example. 
\end{abstract}

%%%%%%%%% BODY TEXT

\section{Introduction}

 Often, you can get a sense of human activity in a movie by listening to the dialogue alone. For example, the sentence \textit{‘Look at that spot over there’}, is an indication that somebody is pointing at something. Similarly, the words \textit{‘Hello, thanks for calling’}, is a good indication that somebody is speaking on the phone. Could this be a valuable source of information for learning good action recognition models?

Obtaining large scale human labelled video datasets to train models
for visual action recognition is a notoriously challenging task.
While large datasets, such as Kinetics~\cite{kay2017kinetics} or
Moments in Time~\cite{monfort2019moments} consisting of individual
short clips (e.g.\ 10s) are now available, these datasets come at
formidable human cost and effort. Furthermore, many such datasets
suffer from heavily skewed distributions with long tails -- i.e.\ it is difficult to obtain
manual labels for \textit{rare} or \textit{infrequent}
actions~\cite{gu2018ava}.

Recently, a number of works have creatively identified \textit{certain domains} of videos, such as
\textit{narrated instructional videos}~\cite{miech2019howto100m,tang2019coin,zhou2018towards} and
 \textit{lifestyle vlogs}~\cite{ignat2019identifying,fouhey2018lifestyle} that are available in huge numbers
(e.g.\ on YouTube) and often contain narration with the
 explicit intention of explaining the visual content on screen.  In
 these video domains, there is a direct link between the action being
 performed, and the speech accompanying the video -- though this link, and the visual 
supervision it provides,  can be quite weak and `noisy' 
as the speech may refer to previous or forthcoming visual 
events, or be about something else entirely~\cite{miech2019howto100m}.

In this paper we explore a complementary link between speech and
actions in the more general domain of movies and TV shows (not
restricted to instructional videos and vlogs). We ask: is it possible
given only a speech sentence  to predict whether an action is
happening,  and, if so, what the action is?  While it appears that in
some cases the speech is correlated with action -- \textit{`Raise your glasses to \ldots'}, in the more general domain of movies and TV shows it is \textit{more} likely that the speech is completely uncorrelated with the action -- \textit{`How is your day going?'}. Hence in this work, we \textit{explicitly} learn to identify when the speech is discriminative. While the supervision we obtain from the speech--action correlation is still noisy, we show that at scale it can provide sufficient weak supervision to train visual classifiers
(see~Fig.~\ref{fig:teaser}).

 Luckily, we have a large amount of literary content at our disposal to learn this correlation between speech and actions. Screenplays can be found for hundreds of movies and TV shows and contain rich descriptions of the identities of people, their actions and interactions with one another and their dialogue. Early work has attempted to \textit{align} these screenplays to the videos themselves, and use that as a source of weak  supervision~\cite{bojanowski2013finding,duchenne2009automatic,laptev2008learning,marszalek2009actions}. However, this is challenging due to the lack of explicit correspondence between scene elements in video and their textual descriptions in screenplays~\cite{bojanowski2013finding}, and notwithstanding alignment quality, is also fundamentally limited in scale to the amount of aligned movie screenplays available.
Instead we learn from \textit{unaligned} movie screenplays. We \textit{first} learn the correlation between speech and actions from written material \textit{alone} and use this to train a \texttt{Speech2Action} classifier. This classifier is then applied to the speech in an unlabelled, unaligned set of videos to obtain
visual samples corresponding to the actions confidently predicted from the speech (Fig.~\ref{fig:teaser}).
In this manner, the correlations can provide us with an effectively infinite source of weak training data, since the audio is freely available with movies.

Concretely, we make the following four contributions: (i) We train a \texttt{Speech2Action} model from literary screenplays, and show that it is possible to predict certain actions from transcribed speech \textit{alone} without the need for any manual labelling; (ii) We apply the \texttt{Speech2Action} model to a large unlabelled corpus of videos to obtain weak labels for video clips from the speech alone; (iii) We demonstrate that an action classifier trained with these weak labels achieves state of the art results for action classification when fine-tuned on standard benchmarks compared to other weakly supervised/domain transfer methods; (iv) Finally, and more interestingly, we evaluate the action classifier trained only on these weak labels with \textit{no} fine-tuning on the mid and tail classes from  the AVA dataset~\cite{gu2018ava} in the zero-shot and few-shot setting, and show a large boost over fully supervised performance for some classes without using a \textit{single} manually labelled example.
\vspace{0.1em}
\section{Related Works}
\noindent\textbf{Aligning Screenplays to Movies:}
 A number of works have explored the use of screenplays to learn and automatically annotate character identity in TV series~\cite{Everingham06a,naim2016aligning,cour2009learning,sivic2009you,tapaswi2012knock}. Learning human actions from
screenplays has also been attempted~\cite{bojanowski2013finding,duchenne2009automatic,laptev2008learning,marszalek2009actions,Miech17}. Crucially, however, all these works rely on aligning these screenplays to the actual videos themselves, often using the speech (as subtitles) to provide correspondences. However,
as noted by~\cite{bojanowski2013finding}, obtaining supervision for actions in this manner
is challenging due to the lack of explicit correspondence between scene elements in video and their textual descriptions in screenplays.

Apart from the imprecise
temporal localization inferred from subtitles correspondences, a major limitation is that
this method is not scalable to all movies and TV shows, since screenplays with stage directions are
simply not available at the same order of magnitude. Hence previous works have been  limited to a small scale, no more 
than \textit{tens} of movies or a season of a TV series~\cite{bojanowski2013finding,duchenne2009automatic,laptev2008learning,marszalek2009actions,Miech17}. A similar argument can be applied to works that align books to movies~\cite{zhu2015aligning, Tapaswi_2015_CVPR}. In contrast, we propose a method that can exploit the richness of information in a modest number of screenplays, and then be applied to a virtually limitless set of edited video material with no alignment or manual annotation required. \\

\noindent\textbf{Supervision for Action Recognition:} 
The benefits of learning from large scale supervised video datasets for the task of action recognition are well known,
with the introduction of datasets like Kinetics~\cite{kay2017kinetics}
spurring the development of new network architectures yielding
impressive performance
gains, e.g.~\cite{carreira2017quo,xie2018rethinking,tran2018closer,wang2017non,WangL_16a,feichtenhofer2019slowfast}.
However, as described in the introduction, such datasets come with
an exorbitant labelling cost. Some work has attempted to reduce this
labeling effort through heuristics~\cite{zhao2017slac} (although a human
annotator is required to clean up the final labels)  or
by procuring weak labels in the form of accompanying meta data such as hashtags~\cite{ghadiyaram2019large}. 

There has \textit{also} been a recent growing interest in using
\textit{cross-modal supervision} from the audio streams readily available with
videos~\cite{owens2016ambient, zhao2018sound,
arandjelovic2017look, owens2018audio,korbar2018cooperative}. Such methods, however, focus on
\textit{non-speech} audio, e.g.\ `guitar playing', the `thud' of a bouncing ball or the `crash' of
waves at the seaside, rather the transcribed speech. As discussed in the introduction, transcribed speech is used only in certain narrow domains, e.g.\ instruction videos~\cite{miech2019howto100m,tang2019coin,zhou2018towards} and lifestyle vlogs~\cite{ignat2019identifying,fouhey2018lifestyle}, while in contrast to these works, we focus on the domain of movies
and TV shows (where the link
between speech and actions is less explicit). Further, such methods
use most or \textit{all} the speech accompanying a video to learn a better
overall visual embedding, whereas we note that often the speech is
completely uninformative of the action. Hence we \textit{first} learn the
correlation between speech and actions from written material, and then
apply this knowledge to an unlabelled set of videos to obtain video
clips that can be used directly for training. \\

\section{\texttt{Speech2Action} Model} 

In this section we describe the steps in data preparation, data mining
and learning, required to train the {\texttt{Speech2Action}} classifier
from a large scale dataset of screenplays. We then assess its
performance in predicting visual actions from transcribed speech segments.

\begin{table*}[ht]
\centering 
\begin{tabular}{c c c c c c c  } 
\toprule
\# movies &
\# scene descriptions &
\# speech segs &
\# sentences &
\# words &
\# unique words &
\# genres \\ 
 \midrule
1,070 &
539,827 &
595,227 &
2,570,993 &
21,364,357 &
590,959 &
22 \\ 
\bottomrule
\end{tabular}
\caption{\textbf{Statistics of the IMSDb dataset of movie screenplays.} This dataset is used to learn the correlation between speech and verbs. We use 850 screenplays for training and 220 for validation. Statistics for sentences and words are from the entire text of the screenplays. }
\label{table:imsdb} 
\end{table*}

\begin{figure*}
\centering 
\begin{scriptsize}
\begin{tabular}{l l l l l l } 
\toprule
 \cellcolor{salmon}&   \cellcolor{salmon}Hello, it's me. & \cellcolor{mistyrose}&    \cellcolor{mistyrose}  One more kiss & \cellcolor{lemonchiffon} &\cellcolor{lemonchiffon}To us        \\ 
 \cellcolor{salmon}&   \cellcolor{salmon}May I have the number for Dr George Shannan & \cellcolor{mistyrose}&  \cellcolor{mistyrose}    Give me a kiss    &  \cellcolor{lemonchiffon} &\cellcolor{lemonchiffon}Raise your glasses to Charlie      \\
\cellcolor{salmon}phone &   \cellcolor{salmon}Honey I asked you not to call unless what why &  \cellcolor{mistyrose} kiss &    \cellcolor{mistyrose}  Good night my darling     &  \cellcolor{lemonchiffon}drink &\cellcolor{lemonchiffon}Heres a toast  \\
\cellcolor{salmon} &   \cellcolor{salmon}hey, it's me & \cellcolor{mistyrose}&     \cellcolor{mistyrose}   I love you my darling     & \cellcolor{lemonchiffon} &\cellcolor{lemonchiffon}You want some water    \\
 \cellcolor{salmon}&   \cellcolor{salmon}Hello, it's me. & \cellcolor{mistyrose}&  \cellcolor{mistyrose}  Noone had ever kissed me there before       & \cellcolor{lemonchiffon} &\cellcolor{lemonchiffon}Drink deep and live       \\
 \cellcolor{salmon} &   \cellcolor{salmon}Hello? & \cellcolor{mistyrose}&    \cellcolor{mistyrose} Goodnight angel my sweet boy &\cellcolor{lemonchiffon}  &\cellcolor{lemonchiffon}Drink up its party time        \\

\cellcolor{lavenderblue} &   \cellcolor{lavenderblue}Shes a beautiful dancer & \cellcolor{azure}&     \cellcolor{azure} So well drop Rudy off at the  bus &\cellcolor{pistachio} &     \cellcolor{pistachio}Officer Van Dorn is right down that hall           \\ 
\cellcolor{lavenderblue} &\cellcolor{lavenderblue}Waddaya say you wanna dance & \cellcolor{azure}&    \cellcolor{azure}  Ill drive her&\cellcolor{pistachio}&\cellcolor{pistachio}OK Print that one             \\
\cellcolor{lavenderblue}dance &   \cellcolor{lavenderblue}Come on Ill take a break and well all dance & \cellcolor{azure} drive & \cellcolor{azure} just parking it out of the way&\cellcolor{pistachio} point &\cellcolor{pistachio}the Metroplitan Museum of Art is right there       \\
\cellcolor{lavenderblue} &\cellcolor{lavenderblue}Ladies and Gentlemen the first dance &\cellcolor{azure}&    \cellcolor{azure}   all you have to do is drop  me off at the bank&\cellcolor{pistachio}& \cellcolor{pistachio}Over there        \\
\cellcolor{lavenderblue}&\cellcolor{lavenderblue}Excuse me would you care for this  dance &\cellcolor{azure}& \cellcolor{azure}Wait down the road&\cellcolor{pistachio}&\cellcolor{pistachio}And her              \\
 \cellcolor{lavenderblue} & \cellcolor{lavenderblue}Hattie do you still dance &\cellcolor{azure}&\cellcolor{azure} He drove around for a long long time driving&\cellcolor{pistachio}&\cellcolor{pistachio}The one with the black spot                \\
\bottomrule
\end{tabular}
\end{scriptsize}
\caption{\textbf{Examples of the top ranked speech samples for six verb categories.} Each block shows the action 
verb on the left, and the speech samples on the right. All speech segments are from the validation set of the IMSDb dataset of movie screenplays.}
\label{table:speechexamples} 
\end{figure*}

\subsection{The IMSDb Dataset}
Movie screenplays are a rich source of data that contain both stage directions (\textit{`Andrew walked over to open the door’}) and the dialogues spoken by the characters (\textit{`Please come in'}). Since stage directions often contain described actions, we use the co-occurrence of dialogue and stage directions in screenplays to learn the relationship between `actions’ and dialogue (see Fig.~\ref{fig:teaser}). In this work, we use a corpus of screenplays extracted from IMSDb (\url{www.imsdb.com}). In order to get a wide variety of different actions (`push' and `kick' as well as `kiss' and `hug') we use screenplays covering a range of different genres\footnote{Action, Adventure, Animation, Biography, Comedy, Crime, Drama, Family, Fantasy, Film-Noir, History, Horror, Music, Musical, Mystery, Romance, Sci-Fi, Short, Sport, Thriller, War, Western}. In total our dataset consists of 1,070 movie screenplays (statistics of the dataset can be seen in Table \ref{table:imsdb}). We henceforth refer to this dataset as the IMSDb dataset. \\
\noindent\textbf{Screenplay Parsing:}
While screenplays (generally) follow a standardized format for their parts (e.g., stage direction, dialogue, location, timing information etc.), they can be challenging to parse due to discrepancies in layout and format. We follow the grammar created by Winer et al.~\cite{winer2017automated} which is based on `The Hollywood Standard'~\cite{riley2009hollywood}, to parse the scripts and separate out various screenplay elements. The grammar provided by~\cite{winer2017automated} parses scripts into the following four different elements, (1) Shot Headings, (2) Stage Directions (which contain mention of actions), (3)~Dialogue and (4) Transitions. More details are provided in Sec. \ref{supp:scripts} of the Appendix. 

In this work we extract only (2) Stage Directions and (3) Dialogue. We extract over 500K stage directions and over 500K dialogue utterances (see Table~\ref{table:imsdb}). It is important to note that since screenplay parsing is done using an automatic method, and sometimes hand-typed screenplays follow completely non-standard formats, this extraction is not perfect. A quick manual inspection of 100 randomly extracted dialogues shows that around 85\% of these are actually dialogue, with the rest being stage directions that have been wrongly labelled as dialogue. \\
\noindent\textbf{Verb Mining the Stage Directions:}
Not all actions will be correlated with speech -- e.g.\ actions like `sitting’ and `standing’ are difficult to distinguish based on speech alone, since they occur commonly with all types of speech. Hence our first endeavour is to automatically determine verbs rendered `discriminative' by speech alone. For this we use the IMSDb dataset described above.  
We first take all the stage directions in the dataset, and break up each sentence into clean word tokens (devoid of punctuation). We then determine the part of speech (PoS) tag for each word using the NLTK toolkit~\cite{loper2002nltk} and obtain a list of all the verbs present. Verbs occurring fewer than 50 times (includes many spelling mistakes) or those occurring too frequently, i.e.\  the top 100 most frequent verbs (these are stop words like `be' etc.) are removed.
For each verb, we then group together all the conjugations and word forms for a particular word stem 
(e.g.\ the stem \textit{run} can appear in many different forms -- running, ran, runs etc.), using the manually created verb conjugations list from the UPenn XTag project\footnote{\url{http://www.cis.upenn.edu/~xtag/}}. All such verb classes are then used in training a BERT-based speech to action classifier, described next.

\subsection{BERT-based Speech Classifier}
Each stage direction is then parsed for verbs belonging to the verb classes identified above. We obtain \textit{paired} speech-action data using proximity in the movie screenplays as a clue. Hence, the nearest speech segment to the stage direction (as illustrated in Fig.~\ref{fig:teaser}) is assigned a label for every verb in the stage direction (more examples in the Appendix, Fig. \ref{fig:screenplays}). This gives us a dataset of speech sentences matched to verb labels. As expected, this is a very noisy dataset. Often, the speech has no correlation with the verb class it is assigned to, and the same speech segment can be assigned to many different verb classes.
To learn the correlation between speech and action, we train a classifier with 850 movies and use the remaining ones for validation. The classifier used is a pretrained BERT~\cite{devlin2018bert} model with an additional classification layer, finetuned on the dataset of speech paired with weak `action' labels. Exact model details are described below. \\
\noindent\textbf{Implementation Details:} 
The model used is BERT-Large Cased with Whole-Word Masking (L=24, H=1024,
A=16, Total Parameters=340M)~\cite{devlin2018bert} pretrained only on English data (BooksCorpus (800M words, \cite{zhu2015aligning}) and the Wikipedia corpus (2,500M words)), since the IMSDb dataset consists only of movie screenplays in English\footnote{The model can be found here: \url{https://github.com/google-research/bert}}. We use WordPiece embeddings~\cite{wu2016google} with a $30,000$ token vocabulary. The first
token of every sequence is always a special classification token ([CLS]). We use the final hidden vector $C \in \mathbb{R}^H$ corresponding to the first
input token ([CLS]) as the aggregate representation. The only new parameters introduced during fine-tuning are classification layer weights $ W \in \mathbb{R}^{K\times H} $ where K is the number of classes. We use the standard cross-entropy loss with $C$ and $W$, i.e., $\log(\text{softmax}(W^TC)$).
We use a batch size of $32$ and finetune the model end-to-end on the IMSDb dataset for 100,000 iterations using the Adam solver with a learning rate of $5\times 10^5$. \\
\noindent\textbf{Results:} \label{sec:bert-results}
We evaluate the performance of our model on the $220$ movie screenplays in the val set. We plot the precision-recall curves using the softmax scores obtained from the \texttt{Speech2Action} model (Fig. \ref{fig:PR} in the Appendix). Only those verbs that achieve an average precision (AP) higher than 0.01 are inferred to be correlated with speech. The highest performing verb classes are `phone', `open' and `run', whereas verb classes like `fishing' and `dig' achieve a very low average precision. We finally conclude that there is a strong correlation for 18 verb classes.\footnote{The verb classes are: `open', `phone', `kiss', `hug', `push', `point', `dance', `drink', `run', `count', `cook', `shoot', `drive', `enter', `fall', `follow', `hit', `eat'. }
Qualitative examples of the most confident predictions (using softmax score as a measure of confidence) for 6 verb classes can be seen in Fig.~\ref{table:speechexamples}.
We note here that we have learnt the correlation between action verb and speech from the movie screenplays using a purely data-driven method. The key assumption is that if there is a \textit{consistent} trend of a verb appearing in the screenplays before or after a speech segment, and our model is able to exploit this trend to minimise a classification objective, we infer that the speech is correlated with the action verb. Because the evaluation is performed purely on the basis of the proximity of speech to verb class in the stage direction of the movie screenplay, it is \textit{not} a perfect ground truth indication of whether an action will actually be performed in a \textit{video} (which is impossible to say only from the movie scripts). We use the stage directions in this case as \textit{pseudo} ground truth, i.e.\ if the stage direction contains an action and the actor then says a particular sentence, we infer that these two must be related.  As a sanity check, we also manually annotate some videos in order to better assess the performance of the \texttt{Speech2Action} model. This is described in Sec. \ref{sec:manual-verif}.

\section{Mining Videos for Action Recognition} 

Now that we have learned the \texttt{Speech2Action} model to map from transcribed speech to actions (from \textit{text} alone), in this section we demonstrate how this can be applied to video.
We use the model to automatically mine video examples from large, unlabelled corpora (the corpus is described in Sec.~\ref{sec:unlab-data}), and assign them with weak labels from the \texttt{Speech2Action} model prediction. Armed with this weakly labelled data, we then train models directly for the downstream task of visual action recognition.
Detailed training and evaluation protocols for the mining are described in the following sections. 
\subsection{Unlabelled Data}\label{sec:unlab-data}
In this work, we apply the \texttt{Speech2Action} model to a large internal corpus of movies and TV shows. 
The corpus consists of $222,855$ movies and TV show episodes. For these videos, we use the closed captions (note that this can be obtained from the audio track directly using automatic speech recognition). The total number of closed captions for this corpus is $188,210,008$, which after dividing into sentences gives us a total of $390,791,653$ (almost 400M) sentences. While we use this corpus in our work, we would like to stress here that there is no correlation between the text data used to train the \texttt{Speech2Action} model and this unlabelled corpus (other than both belonging to the movie domain), and such a model can be applied to any other corpus of unlabelled, edited film material. 
\subsection{Obtaining Weak Labels}
In this section, we describe how we obtain weak action labels for short clips from the speech alone. We do this in two ways, (i) using the \texttt{Speech2Action} model, and (ii) using a simple keyword spotting baseline described below. 
\subsubsection{Using \texttt{Speech2Action}} \label{sec:weak}
The \texttt{Speech2Action} model is applied to a single sentence of speech, and the prediction is used as a weak label if the confidence (softmax score) is above a certain threshold. The threshold is obtained by taking the confidence value at a precision of 0.3 on the IMSDb validation set, with some manual adjustments for the classes of `phone', `run' and `open' (since these classes have a much higher recall, we increase the threshold in order to prevent a huge imbalance of retrieved samples). More details are provided in the Appendix, Sec.~\ref{supp:PR}. We then extract the visual frames for a 10 second clip centered around the midpoint of the timeframe spanned by the caption, and assign the \texttt{Speech2Action} label as the weak label for the clip. Ultimately, we successfully end up mining $837,334$ video clips for 18 action classes. While this is a low yield,
we still end up with a large number of mined clips, greater than the manually labelled Kinetics dataset~\cite{kay2017kinetics} (600K). 

\begin{figure}[h]
\begin{center}
\includegraphics[width=0.4\textwidth]{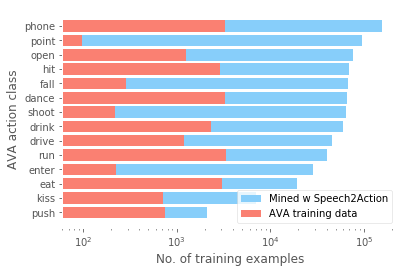}
\end{center}
   \caption{\textbf{Distribution of training clips mined using
   \texttt{Speech2Action}}. We compare the distribution of mined clips to the number of samples in the AVA training set. Although the mined clips are noisy, we are able to obtain far more, in some cases up to \textit{two} orders of magnitude more training data (note the \textbf{log scale} in the x-axis). }
\label{fig:trainingdist}
\end{figure}

\begin{figure*}[h]
\begin{center}
\includegraphics[width=1\textwidth]{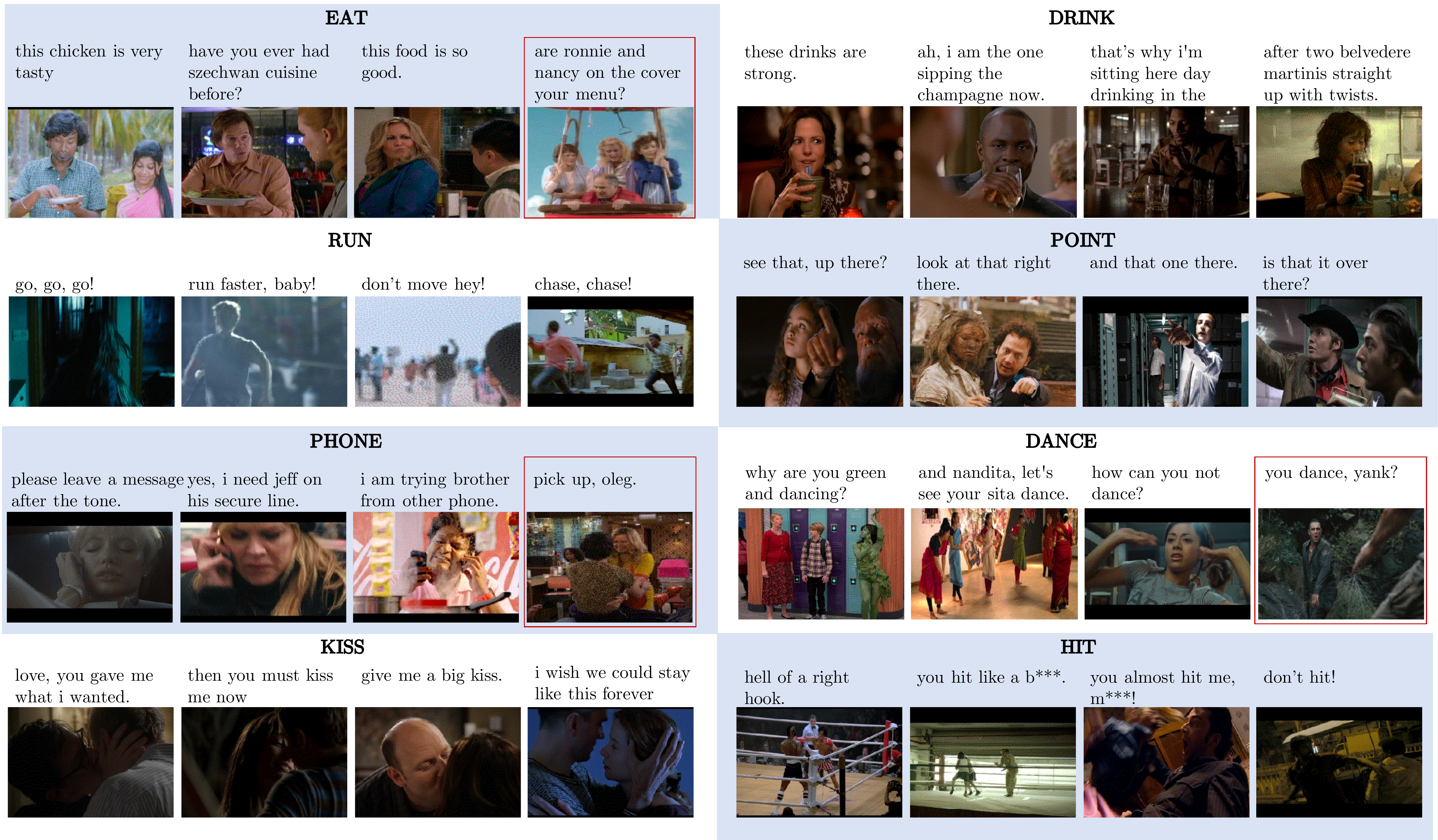}
\end{center}
   \caption{\textbf{Examples of clips mined automatically using the \texttt{Speech2Action} model applied to \textit{speech alone} for 8 AVA classes.} We show only a single frame from each video. Note the diversity in background, actor and view point. We show false positives for eat, phone and dance (last in each row, enclosed in a red box). Expletives are censored. More examples are provided in the Appendix.}
\label{fig:mining}
\end{figure*}

We also discover that the verb classes that have high correlation with
speech in the IMSDb dataset tend to be \textit{infrequent} or
\textit{rare} actions in other datasets~\cite{gu2018ava} -- as shown in Fig.~\ref{fig:trainingdist}, we obtain two orders of
magnitude more data for certain classes in the AVA training set~\cite{gu2018ava}.  
Qualitative examples of mined
video clips with action labels can be seen in
Fig.~\ref{fig:mining}. Note how we are able to retrieve clips with a
wide variety in background and actor, simply from the speech alone. Refer to Fig.~\ref{fig:mining_more} in the Appendix for more examples showing diversity in objects and viewpoint.

\subsubsection{Using a Keyword Spotting Baseline}
In order to validate the efficacy of our \texttt{Speech2Action} model trained on movie screenplays, we also compare to a simple keyword spotting baseline. This involves searching for the action verb in the speech directly -- a speech segment like \textit{`Will you eat now?'} is directly assigned the label `eat'. This itself is a very powerful baseline, e.g.\ speech segments such as \textit{`Will you dance with me'}, are strongly indicative of the action `dance'. To implement this baseline, we search for the presence of the action verb (or its conjugations) in the speech segment directly, and if the verb is present in the speech, we assign the action label to the video clip directly. The fallacy of this method is that there is no distinction between the different semantic meanings of a verb, e.g.\ the speech segment \textit{`You've missed the point entirely'} will be weakly labelled with the verb `point' using this baseline, which is indicative of a different semantic meaning to the physical action of `pointing'. Hence as we show in the results, this baseline performs poorly compared to our \texttt{Speech2Action} mining method (Tables~\ref{tab:per-class} and~\ref{tab:sota-hmdb}). More examples of speech labelled using this keyword spotting baseline can be seen in Table \ref{table:baselineexamples} in the Appendix.
\\
\subsubsection{Manual Evaluation of \texttt{Speech2Action}} \label{sec:manual-verif}
\begin{table}
\centering 
\scriptsize{
\begin{tabular}{c c c c c c c c c c } 
\toprule
dance &
phone &
kiss &
drive &
eat &
drink & run & point & hit & shoot \\ 
\midrule
42 & 68 & 18 & 41 & 27 & 51 & 83 &52 & 18 & 27\\

 \bottomrule
\end{tabular}
}
\caption{\textbf{Number of true positives for 100 randomly retrieved samples for $10$ classes.} These estimates are obtained through manual inspection of video clips that are labelled with \texttt{Speech2Action}. While the true positive rate for some classes is low, the other samples still contain valuable information for the classifier. For example, although there are only 18 true samples of `kiss', many of the other videos have two people with their lips very close together, or even if they are not `eating' strictly, many times they are holding food in their hands.}
\label{tab:manual} 
\end{table}
We now assess the performance of \texttt{Speech2Action} applied to videos. Given a speech segment, we check whether a prediction made by the model on the speech translates to the action being performed visually in the frames aligned to the speech. To assess this, we do a manual inspection of a random set of 100 retrieved video clips for $10$ of the verb classes, and report the true positive rate (number of clips for which the action is visible) in Table~\ref{tab:manual}. We find that a surprising number of samples actually contain the action during the time frame of $10$ seconds, with some classes noisier than others. The high purity of the classes `run' and `phone' can be explained by the higher thresholds used for mining, as explained in Sec.~\ref{sec:weak}. Common sources of false positives are actions performed off screen, or actions performed at a temporal offset (either much before or much after) the speech segment. We note that at no point do we ever actually use any of the manual labels for training, these are purely for evaluation and as a sanity check.

\section{Action Classification}
Now that we have described our method to obtain weakly labelled training data, we train 
a video classifier with the S3D-G~\cite{xie2018rethinking} backbone  on these noisy samples for the task of action recognition. We first detail the training and testing protocols, and then describe the datasets used in this work.

\subsection{Evaluation Protocol}
We evaluate our video classifier for the task of action classification in the following two ways: \\
First, we follow the typical procedure adopted in the video understanding literature~\cite{carreira2017quo}: pre-training on a large corpus of videos weakly labelled using our \texttt{Speech2Action} model, followed by fine-tuning on the training split of a labeled target
dataset (‘test bed’). After training, we evaluate the performance on the test set of the target dataset. In this work we use HMDB-51~\cite{kuehne2011hmdb}, and compare to other state of the art methods on this dataset. We also provide results for the UCF101 dataset~\cite{soomro2012ucf101} in Sec.~\ref{supp:ucf} of the Appendix. \\
Second, and perhaps more interestingly, we apply our method by training a video classifier on the mined video clips for some action classes, and evaluating it \textit{directly} on the test samples of \textit{rare} action classes in the target dataset (in this case we use the AVA dataset~\cite{gu2018ava}). Note: At this point we also manually verified that there is no overlap between the movies in the IMSDb dataset and the AVA dataset (not surprising since AVA movies are older and more obscure – these are movies that are freely available on YouTube). Here not a single manually labelled training example is used, since there is no finetuning (we henceforth refer to this as zero-shot\footnote{In order to avoid confusion with the strict meaning of this term, we clarify that in this work we use it to refer to the case where not a single \textit{manually labelled} example is available for a particular class. We do however train on multiple weakly labelled examples.}). We also report performance for the few-shot learning scenario, where we fine-tune our model on a \textit{small} number of labelled examples. We note that in this case, we can only evaluate on the classes that directly overlap with the verb classes in the IMSDb dataset.
\subsection{Datasets and Experimental Details}\label{sec:data}
\noindent\textbf{HMDB51:} HMDB51~\cite{kuehne2011hmdb} contains 6,766 realistic
and varied video clips from 51 action classes. Evaluation is
performed using average classification accuracy over three
train/test splits from~\cite{idrees2017thumos}, each with 3,570 train and 1,530 test videos. \\
\noindent\textbf{AVA:} 
The AVA dataset~\cite{gu2018ava} is collected by exhaustively manually annotating videos and exhibits a strong imbalance in the number of examples between the common and rare classes. Eg. a common action, like `stand', has 160K training and 43K test examples, compared to `drive' (1.18K train and 561 test) and `point' (only 96 train and 32 test). As a result, methods relying on full supervision struggle on the categories in the middle and the end of the tail. We evaluate on the $14$ AVA classes that overlap with the classes present in the IMDSDb dataset (all from the middle and tail). While the dataset is originally a detection dataset, we repurpose it simply for the task of action classification, by assigning each frame the union of labels from all bounding box annotations. We then train and test on samples from these $14$ action classes, reporting per-class average precision (AP).

\noindent\textbf{Implementation Details:} 
We train the S3D with gating (S3D-G)~\cite{xie2018rethinking} model as our visual classifier. Following~\cite{xie2018rethinking}, we densely sample 64 frames from a video, resize input frames to $256 \times 256$ and then take random crops of size $224 \times 224$ during training. During evaluation, we use all frames and take $224 \times 224$ center crops from the resized frames. Our models are implemented with TensorFlow and optimized with a vanilla synchronous SGD algorithm with momentum of 0.9. For models trained from scratch, we train for 150K iterations with a learning rate schedule of $10^2$, $10^3$ and $10^4$ dropping after 80K and 100K iterations, and for finetuning we train for 60K iterations using a learning rate of $10^2$. \\ 
\noindent\textbf{Loss functions for training:}
We try both the softmax cross-entropy and per-class sigmoid loss, and find that the performance was
relatively stable with both choices.

\subsection{Results} 

\begin{table}[ht]
\centering 
\footnotesize{
\begin{tabular}{l l l  l  r}
\toprule
Method & Architecture  & Pre-training & Acc. \\

\midrule
Shuffle\&Learn~\cite{misra2016shuffle}$\star$ & S3D-G (RGB) & UCF101$\dagger$~\cite{soomro2012ucf101}& 35.8\\
OPN \cite{lee2017unsupervised}  & VGG-M-2048  & UCF101$\dagger$~\cite{soomro2012ucf101}  & 23.8 \\
ClipOrder \cite{xu2019self} & R(2+1)D  & UCF101$\dagger$~\cite{soomro2012ucf101}   & 30.9 \\
Wang et al. \cite{wang2019self} & C3D  & Kinetics$\dagger$~\cite{soomro2012ucf101}   & 33.4 \\
3DRotNet \cite{jing2018self}$\star$ & S3D-G (RGB)  & Kinetics$\dagger$   &  40.0 \\
DPC \cite{han2019video} & 3DResNet18 & Kinetics$\dagger$ & 35.7 \\
CBT \cite{sun2019contrastive} & S3D-G (RGB) & Kinetics$\dagger$   & 44.6 \\
\midrule 
DisInit (RGB) \cite{girdhar2019distinit} & R(2+1)D-18 \cite{tran2018closer} & Kinetics$**$  & 54.8 \\  
Korbar et al \cite{korbar2018cooperative} & I3D (RGB) & Kinetics$\dagger$  & 53.0 \\  
\midrule 
- & S3D-G (RGB) & Scratch& 41.2 \\ 
Ours & S3D-G (RGB)  & \texttt{KSB-mined}  & 46.0 \\
Ours & S3D-G (RGB)  & \texttt{S2A-mined}  & \textbf{58.1} \\
\midrule
Supervised pretraining & S3D-G (RGB)  & ImageNet & 54.7  \\
Supervised pretraining & S3D-G (RGB)  & Kinetics & 72.3  \\
\bottomrule
\end{tabular}}
\caption{{\textbf{Action classification results on HMDB51.} Pre-training on videos labelled with \texttt{Speech2Action} leads to a 17\% improvement over training from scratch and also outperforms previous self-supervised and weakly supervised works.
\textbf{KSB-mined:} video clips mined using the keyword spotting baseline.
\textbf{S2A-mined:} video clips mined using the \texttt{Speech2Action} model. $\dagger$videos without labels. **videos with labels distilled from ImageNet.  When comparing to~\cite{korbar2018cooperative}, we report the number achieved by their I3D (RGB only) model which is the closest to our architecture. For $\star$, we report the reimplementations by \cite{sun2019contrastive} using the S3D-G model (same as ours). For the rest, we report performance directly from the original papers.}}
\label{tab:sota-hmdb}
\end{table} 
\begin{table*}[ht]
\centering 
\footnotesize{
\begin{tabular}{c c c c c c c c c c c c c c c c} 
\toprule
\multicolumn{1}{c}{\textbf{Data}}&
\multicolumn{14}{c}{\textbf{Per-Class AP}}  \\
                              & drive & phone & kiss & dance & eat & drink & run & point & open & hit & shoot & push & hug & enter \\ 
\midrule
AVA \tiny{(fully supervised)} & 0.63  & 0.54  & 0.22 & 0.46  &0.67 & 0.27  & 0.66 & 0.02 & 0.49 & 0.62 & 0.08 & 0.09 & 0.29 &0.14 \\
\midrule
KS-baseline $\dagger$         & 0.67 & 0.20   & 0.12& 0.53 & 0.67 & 0.18   & 0.37 & 0.00 & 0.33 & 0.47 & 0.05 & 0.03 & 0.10 & 0.02\\
 \texttt{S2A-mined} \tiny{(zero-shot)} & \cellcolor{mistyrose}{0.83} & \cellcolor{mistyrose}{0.79} & 0.13 & \cellcolor{mistyrose}{0.55} & \cellcolor{mistyrose}{0.68} & \cellcolor{mistyrose}{0.30} & 0.63 &  \cellcolor{mistyrose}{0.04}& \cellcolor{mistyrose}{0.52} & 0.54 & \cellcolor{mistyrose}{0.18} & 0.04 & 0.07 & 0.04\\
\texttt{S2A-mined} + AVA      & 0.84  & 0.83  & 0.18 & 0.56 & 0.75 & 0.40 & 0.74 & 0.05 & 0.56 & 0.64 & 0.23 & 0.07 & 0.17 &0.04 \\
\midrule
  AVA \tiny{(few-shot)}-20 & 0.82 & 0.83 & 0.22 & 0.55 & 0.69 & 0.33 & 0.64& 0.04& 0.51 & 0.59 & 0.20 & 0.06 & 0.19 & 0.13\\
  AVA \tiny{(few-shot)}-50 & 0.82 & 0.85 & 0.26 & 0.56 & 0.70 & 0.37 & 0.69& 0.04& 0.52 & 0.65 & 0.21 & 0.06 & 0.19 & 0.15\\
  AVA \tiny{(few-shot)}-100 & 0.84& 0.86 & 0.30 & \textbf{0.58} & 0.71 & 0.39 & \textbf{0.75} & \textbf{0.05} & 0.58 & 0.73 & 0.25 & \textbf{0.13} & 0.27 & 0.15 \\
AVA (all) & \textbf{0.86} & \textbf{0.89} & \textbf{0.34} & \textbf{0.58} & \textbf{0.78} & \textbf{0.42} & \textbf{0.75} & 0.03 & \textbf{0.65} & \textbf{0.72} & \textbf{0.26} & \textbf{0.13} & \textbf{0.36} & \textbf{0.16}\\
 \bottomrule
\end{tabular}
}
\caption{{\textbf{Per-class average precision for $14$ AVA mid and tail classes.} These actions occur \textit{rarely}, and hence are harder to get manual supervision for. For 8 of the 14 classes, we exceed fully supervised performance without a single manually labelled training example (highlighted in pink, best viewed in colour). \texttt{S2A-mined:} Video clips mined using \texttt{Speech2Action}. $\dagger$ Keyword spotting baseline. First $4$ rows: models are trained from scratch. Last $4$ rows: we pre-train on video clips mined using \texttt{Speech2Action}. }}
\label{tab:per-class} 
\end{table*}

\begin{figure*}[h]
\begin{center}
\includegraphics[width=1\textwidth]{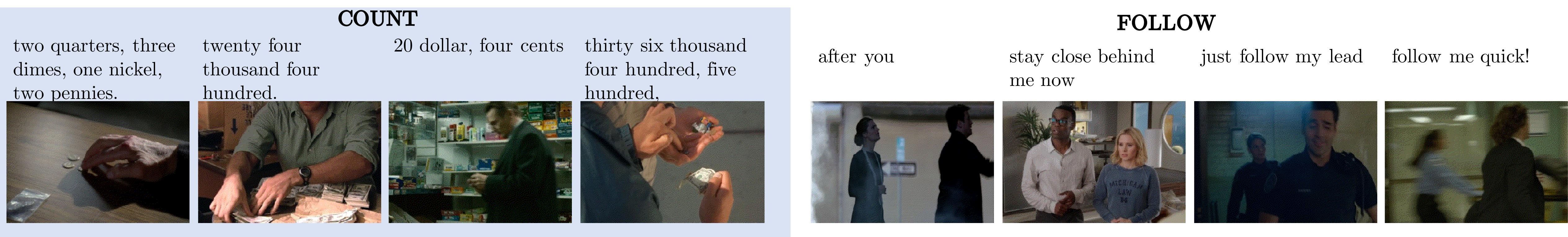}
\end{center}
   \caption{\textbf{Examples of clips mined for more abstract actions.} These are actions that are not present in standard datasets like HMDB51 or AVA, but are quite well correlated with speech. Our method is able to automatically mine clips weakly labelled with these actions from unlabelled data.}
\label{fig:mining-unusual}
\end{figure*}
\noindent\textbf{HMDB51:} The results on HMDB51 can be seen in Table~\ref{tab:sota-hmdb}. Training on videos labelled with \texttt{Speech2Actions} leads to a significant 17\% improvement over from-scratch training. For reference, we also compare to other self-supervised and weakly supervised works (note that these methods differ both in architecture and training objective). We show a 14\% improvement over previous self-supervised works that use \textit{only} video frames (no other modalities). We also compare to Korbar {\it et al.}~\cite{korbar2018cooperative} who pretrain using audio and video synchronisation on AudioSet, DisInit~\cite{girdhar2019distinit}, which distills knowledge from ImageNet into Kinetics videos, and simply pretraining on ImageNet and then inflating 2D convolutions to our S3D-G model~\cite{kay2017kinetics}. 
We improve over these works by 3-4\% -- which is impressive given that the latter two methods rely on access to a large-scale manually labelled image dataset \cite{deng2009imagenet}, whereas ours relies only on 1000 unlabelled movie scripts. Another point of interest (and perhaps an unavoidable side-effect of this stream of self- and weak-supervision) is that while all these previous methods do not use labels, they still pretrain on the Kinetics data, which has been carefully curated to cover a wide diversity of over 600 different actions. In contrast, we mine our training data directly from movies, without the need for any manual labelling or careful curation, and our pretraining data was mined for only $18$ classes. \\
\noindent\textbf{AVA-scratch:} The results on AVA for models trained from scratch with \textit{no} pretraining, can be seen in Table \ref{tab:per-class} (top 4 rows). We compare the following: training with the AVA training examples (Table \ref{tab:per-class}, top row), training only with our mined examples, and training jointly with both. For 8 out of 14 classes, we exceed fully supervised performance without a single AVA training example, in some cases (`drive' and `phone') almost by 20\%.  \\
\noindent\textbf{AVA-finetuned:} We also show results for pre-training on \texttt{Speech2Action} mined clips first, and then fine-tuning on a gradually increasing number of AVA labelled training samples per class (Table \ref{tab:per-class}, bottom $4$ rows). Here we keep all the weights from the fine-tuning, including the classification layer weights, for initialisation, and fine-tune only for a single epoch. With $50$ training samples per class, we exceed fully supervised performance for all classes (except for `hug' and `push') compared to training from scratch. The worst performance is for the class `hug' -- `hug' and `kiss' are often confused, as the speech in both cases tends to be similar -- \textit{'I love you'}. A quick manual inspection shows that most of the clips are wrongly labelled as `kiss', which is why we are only able to mine very few video clips for this class. For completeness, we also pretrain a model with the S2A mined clips (only 14 classes) and then finetune on AVA for \textit{all 60} classes used for evaluation, and get a 40\% overall classification acc.\ vs 38\% with training on AVA alone. \\
\noindent\textbf{Mining Technique:} We also train on clips mined using the keyword spotting baseline (Table \ref{tab:per-class}). For some classes, this baseline itself exceeds fully supervised performance. Our \texttt{Speech2Action} labelling beats this baseline for all classes, indeed the baseline does poorly for classes like `point' and `open' -- verbs which have many semantic meanings, demonstrating that the semantic information learnt from the IMSDb dataset is valuable. However we note here that it is difficult to measure performance quantitatively for the class `point' due to idiosyncrasies in the AVA test set (wrong ground truth labels for very few test samples) and hence we show qualitative examples of mined clips in Fig. \ref{fig:mining}. We note that the baseline comes very close for `dance' and `eat', demonstrating that simple keyword matching on speech can retrieve good training data for these actions. \\
\noindent\textbf{Abstract Actions:} By gathering data directly from the stage directions in movie screenplays, our action labels are post-defined (as in~\cite{fouhey2018lifestyle}). This is unlike the majority of the existing human action datasets that use pre-defined labels~\cite{caba2015activitynet, sigurdsson2016hollywood, gu2018ava, monfort2019moments}. Hence we also manage to mine examples for some unusual or \textit{abstract} actions which are quite well correlated with speech, such as `count' and `follow'. While these are not present in standard action recognition datasets such as HMDB51 or AVA, and hence cannot be evaluated numerically, we show some qualitative examples of these mined videos in Fig. \ref{fig:mining-unusual}.

\section{Conclusion}
We provide a new data-driven approach to obtain weak labels for action recognition, using speech alone. With only a thousand unaligned screenplays as a starting point, we obtain weak labels automatically for a number of rare action classes. 
However, there is a plethora of literary material available online, including plays and books, and exploiting these sources of text may allow us to extend our method to predict other action classes, including composite actions of `verb' and `object'. We also note that \textit{besides} actions, people talk about physical objects, events and scenes -- descriptions of which are also present in screenplays and books. Hence the same principle used here could be applied to mine videos for more general visual content.

\noindent\textbf{Acknowledgments:}
Arsha is supported by a Google PhD Fellowship. We are grateful to Carl Vondrick for early discussions.
\clearpage

{\small
\bibliographystyle{ieee_fullname}

\begin{thebibliography}{10}\itemsep=-1pt

\bibitem{arandjelovic2017look}
Relja Arandjelovic and Andrew Zisserman.
\newblock Look, listen and learn.
\newblock In {\em Proceedings of the IEEE International Conference on Computer
  Vision}, pages 609--617, 2017.

\bibitem{bojanowski2013finding}
Piotr Bojanowski, Francis Bach, Ivan Laptev, Jean Ponce, Cordelia Schmid, and
  Josef Sivic.
\newblock Finding actors and actions in movies.
\newblock In {\em Proceedings of the IEEE international conference on computer
  vision}, pages 2280--2287, 2013.

\bibitem{caba2015activitynet}
Fabian Caba~Heilbron, Victor Escorcia, Bernard Ghanem, and Juan Carlos~Niebles.
\newblock Activitynet: A large-scale video benchmark for human activity
  understanding.
\newblock In {\em Proceedings of the IEEE Conference on Computer Vision and
  Pattern Recognition}, pages 961--970, 2015.

\bibitem{carreira2017quo}
Joao Carreira and Andrew Zisserman.
\newblock Quo vadis, action recognition? a new model and the {Kinetics}
  dataset.
\newblock In {\em proceedings of the IEEE Conference on Computer Vision and
  Pattern Recognition}, pages 6299--6308, 2017.

\bibitem{corro2014werdy}
Luciano~Del Corro, Rainer Gemulla, and Gerhard Weikum.
\newblock Werdy: Recognition and disambiguation of verbs and verb phrases with
  syntactic and semantic pruning.
\newblock 2014.

\bibitem{cour2009learning}
Timothee Cour, Benjamin Sapp, Chris Jordan, and Ben Taskar.
\newblock Learning from ambiguously labeled images.
\newblock In {\em 2009 IEEE Conference on Computer Vision and Pattern
  Recognition}, pages 919--926. IEEE, 2009.

\bibitem{deng2009imagenet}
Jia Deng, Wei Dong, Richard Socher, Li-Jia Li, Kai Li, and Li Fei-Fei.
\newblock Imagenet: A large-scale hierarchical image database.
\newblock In {\em 2009 IEEE conference on computer vision and pattern
  recognition}, pages 248--255. Ieee, 2009.

\bibitem{devlin2018bert}
Jacob Devlin, Ming-Wei Chang, Kenton Lee, and Kristina Toutanova.
\newblock Bert: Pre-training of deep bidirectional transformers for language
  understanding.
\newblock {\em arXiv preprint arXiv:1810.04805}, 2018.

\bibitem{duchenne2009automatic}
Olivier Duchenne, Ivan Laptev, Josef Sivic, Francis Bach, and Jean Ponce.
\newblock Automatic annotation of human actions in video.
\newblock In {\em 2009 IEEE 12th International Conference on Computer Vision},
  pages 1491--1498. IEEE, 2009.

\bibitem{Everingham06a}
Mark Everingham, Josef Sivic, and Andrew Zisserman.
\newblock ``{H}ello! {M}y name is... {Buffy}'' -- automatic naming of
  characters in {TV} video.
\newblock In {\em BMVC}, 2006.

\bibitem{feichtenhofer2019slowfast}
Christoph Feichtenhofer, Haoqi Fan, Jitendra Malik, and Kaiming He.
\newblock Slowfast networks for video recognition.
\newblock In {\em Proceedings of the IEEE International Conference on Computer
  Vision}, pages 6202--6211, 2019.

\bibitem{fouhey2018lifestyle}
David~F Fouhey, Wei-cheng Kuo, Alexei~A Efros, and Jitendra Malik.
\newblock From lifestyle vlogs to everyday interactions.
\newblock In {\em Proceedings of the IEEE Conference on Computer Vision and
  Pattern Recognition}, pages 4991--5000, 2018.

\bibitem{ghadiyaram2019large}
Deepti Ghadiyaram, Du Tran, and Dhruv Mahajan.
\newblock Large-scale weakly-supervised pre-training for video action
  recognition.
\newblock In {\em Proceedings of the IEEE Conference on Computer Vision and
  Pattern Recognition}, pages 12046--12055, 2019.

\bibitem{girdhar2019distinit}
Rohit Girdhar, Du Tran, Lorenzo Torresani, and Deva Ramanan.
\newblock Distinit: Learning video representations without a single labeled
  video.
\newblock {\em ICCV}, 2019.

\bibitem{gu2018ava}
Chunhui Gu, Chen Sun, David~A Ross, Carl Vondrick, Caroline Pantofaru, Yeqing
  Li, Sudheendra Vijayanarasimhan, George Toderici, Susanna Ricco, Rahul
  Sukthankar, Cordelia Schmid, and Jitendra Malik.
\newblock {AVA}: A video dataset of spatio-temporally localized atomic visual
  actions.
\newblock In {\em Proceedings of the IEEE Conference on Computer Vision and
  Pattern Recognition}, pages 6047--6056, 2018.

\bibitem{han2019video}
Tengda Han, Weidi Xie, and Andrew Zisserman.
\newblock Video representation learning by dense predictive coding.
\newblock In {\em Proceedings of the IEEE International Conference on Computer
  Vision Workshops}, 2019.

\bibitem{idrees2017thumos}
Haroon Idrees, Amir~R Zamir, Yu-Gang Jiang, Alex Gorban, Ivan Laptev, Rahul
  Sukthankar, and Mubarak Shah.
\newblock The {THUMOS} challenge on action recognition for videos “in the
  wild”.
\newblock {\em Computer Vision and Image Understanding}, 155:1--23, 2017.

\bibitem{ignat2019identifying}
Oana Ignat, Laura Burdick, Jia Deng, and Rada Mihalcea.
\newblock Identifying visible actions in lifestyle vlogs.
\newblock {\em arXiv preprint arXiv:1906.04236}, 2019.

\bibitem{jing2018self}
Longlong Jing and Yingli Tian.
\newblock Self-supervised spatiotemporal feature learning by video geometric
  transformations.
\newblock {\em arXiv preprint arXiv:1811.11387}, 2018.

\bibitem{kay2017kinetics}
W. Kay, J. Carreira, K. Simonyan, B. Zhang, C. Hillier, S. Vijayanarasimhan, F.
  Viola, T. Green, T. Back, P. Natsev, M. Suleyman, and A. Zisserman.
\newblock The {Kinetics} human action video dataset.
\newblock {\em CoRR}, abs/1705.06950, 2017.

\bibitem{korbar2018cooperative}
Bruno Korbar, Du Tran, and Lorenzo Torresani.
\newblock Cooperative learning of audio and video models from self-supervised
  synchronization.
\newblock In {\em Advances in Neural Information Processing Systems}, pages
  7763--7774, 2018.

\bibitem{kuehne2011hmdb}
Hildegard Kuehne, Hueihan Jhuang, Est{\'\i}baliz Garrote, Tomaso Poggio, and
  Thomas Serre.
\newblock Hmdb: a large video database for human motion recognition.
\newblock In {\em 2011 International Conference on Computer Vision}, pages
  2556--2563. IEEE, 2011.

\bibitem{laptev2008learning}
Ivan Laptev, Marcin Marsza{\l}ek, Cordelia Schmid, and Benjamin Rozenfeld.
\newblock Learning realistic human actions from movies.
\newblock In {\em IEEE Conference on Computer Vision \& Pattern Recognition},
  2008.

\bibitem{lee2017unsupervised}
Hsin-Ying Lee, Jia-Bin Huang, Maneesh Singh, and Ming-Hsuan Yang.
\newblock Unsupervised representation learning by sorting sequences.
\newblock In {\em Proceedings of the IEEE International Conference on Computer
  Vision}, pages 667--676, 2017.

\bibitem{loper2002nltk}
Edward Loper and Steven Bird.
\newblock Nltk: the natural language toolkit.
\newblock {\em arXiv preprint cs/0205028}, 2002.

\bibitem{marszalek2009actions}
Marcin Marsza{\l}ek, Ivan Laptev, and Cordelia Schmid.
\newblock Actions in context.
\newblock In {\em CVPR 2009-IEEE Conference on Computer Vision \& Pattern
  Recognition}, pages 2929--2936. IEEE Computer Society, 2009.

\bibitem{Miech17}
Antoine Miech, Jean{-}Baptiste Alayrac, Piotr Bojanowski, Ivan Laptev, and
  Josef Sivic.
\newblock Learning from video and text via large-scale discriminative
  clustering.
\newblock In {\em Proceedings of the IEEE international conference on computer
  vision}, 2017.

\bibitem{miech2019howto100m}
Antoine Miech, Dimitri Zhukov, Jean-Baptiste Alayrac, Makarand Tapaswi, Ivan
  Laptev, and Josef Sivic.
\newblock {HowTo100M: Learning a Text-Video Embedding by Watching Hundred
  Million Narrated Video Clips}.
\newblock In {\em Proceedings of the IEEE international conference on computer
  vision}, 2019.

\bibitem{misra2016shuffle}
Ishan Misra, C~Lawrence Zitnick, and Martial Hebert.
\newblock Shuffle and learn: unsupervised learning using temporal order
  verification.
\newblock In {\em European Conference on Computer Vision}, pages 527--544.
  Springer, 2016.

\bibitem{monfort2019moments}
Mathew Monfort, Alex Andonian, Bolei Zhou, Kandan Ramakrishnan, Sarah~Adel
  Bargal, Yan Yan, Lisa Brown, Quanfu Fan, Dan Gutfreund, Carl Vondrick, et~al.
\newblock Moments in time dataset: one million videos for event understanding.
\newblock {\em IEEE transactions on pattern analysis and machine intelligence},
  2019.

\bibitem{naim2016aligning}
Iftekhar Naim, Abdullah Al~Mamun, Young~Chol Song, Jiebo Luo, Henry Kautz, and
  Daniel Gildea.
\newblock Aligning movies with scripts by exploiting temporal ordering
  constraints.
\newblock In {\em 2016 23rd International Conference on Pattern Recognition
  (ICPR)}, pages 1786--1791. IEEE, 2016.

\bibitem{owens2018audio}
Andrew Owens and Alexei~A Efros.
\newblock Audio-visual scene analysis with self-supervised multisensory
  features.
\newblock In {\em Proceedings of the European Conference on Computer Vision
  (ECCV)}, pages 631--648, 2018.

\bibitem{owens2016ambient}
Andrew Owens, Jiajun Wu, Josh~H McDermott, William~T Freeman, and Antonio
  Torralba.
\newblock Ambient sound provides supervision for visual learning.
\newblock In {\em European conference on computer vision}, pages 801--816.
  Springer, 2016.

\bibitem{riley2009hollywood}
Christopher Riley.
\newblock {\em The Hollywood standard: the complete and authoritative guide to
  script format and style}.
\newblock Michael Wiese Productions, 2009.

\bibitem{sigurdsson2016hollywood}
Gunnar~A Sigurdsson, G{\"u}l Varol, Xiaolong Wang, Ali Farhadi, Ivan Laptev,
  and Abhinav Gupta.
\newblock Hollywood in homes: Crowdsourcing data collection for activity
  understanding.
\newblock In {\em European Conference on Computer Vision}, pages 510--526.
  Springer, 2016.

\bibitem{sivic2009you}
Josef Sivic, Mark Everingham, and Andrew Zisserman.
\newblock “who are you?”-learning person specific classifiers from video.
\newblock In {\em 2009 IEEE Conference on Computer Vision and Pattern
  Recognition}, pages 1145--1152. IEEE, 2009.

\bibitem{soomro2012ucf101}
Khurram Soomro, Amir~Roshan Zamir, and Mubarak Shah.
\newblock Ucf101: A dataset of 101 human actions classes from videos in the
  wild.
\newblock {\em arXiv preprint arXiv:1212.0402}, 2012.

\bibitem{sun2019contrastive}
Chen Sun, Fabien Baradel, Kevin Murphy, and Cordelia Schmid.
\newblock Contrastive bidirectional transformer for temporal representation
  learning.
\newblock {\em arXiv preprint arXiv:1906.05743}, 2019.

\bibitem{tang2019coin}
Yansong Tang, Dajun Ding, Yongming Rao, Yu Zheng, Danyang Zhang, Lili Zhao,
  Jiwen Lu, and Jie Zhou.
\newblock Coin: A large-scale dataset for comprehensive instructional video
  analysis.
\newblock In {\em Proceedings of the IEEE Conference on Computer Vision and
  Pattern Recognition}, pages 1207--1216, 2019.

\bibitem{tapaswi2012knock}
Makarand Tapaswi, Martin B{\"a}uml, and Rainer Stiefelhagen.
\newblock “knock! knock! who is it?” probabilistic person identification in
  tv-series.
\newblock In {\em 2012 IEEE Conference on Computer Vision and Pattern
  Recognition}, pages 2658--2665. IEEE, 2012.

\bibitem{Tapaswi_2015_CVPR}
Makarand Tapaswi, Martin Bauml, and Rainer Stiefelhagen.
\newblock Book2movie: Aligning video scenes with book chapters.
\newblock In {\em The IEEE Conference on Computer Vision and Pattern
  Recognition (CVPR)}, June 2015.

\bibitem{tran2018closer}
Du Tran, Heng Wang, Lorenzo Torresani, Jamie Ray, Yann LeCun, and Manohar
  Paluri.
\newblock A closer look at spatiotemporal convolutions for action recognition.
\newblock In {\em Proceedings of the IEEE conference on Computer Vision and
  Pattern Recognition}, pages 6450--6459, 2018.

\bibitem{wang2019self}
Jiangliu Wang, Jianbo Jiao, Linchao Bao, Shengfeng He, Yunhui Liu, and Wei Liu.
\newblock Self-supervised spatio-temporal representation learning for videos by
  predicting motion and appearance statistics.
\newblock In {\em Proceedings of the IEEE Conference on Computer Vision and
  Pattern Recognition}, pages 4006--4015, 2019.

\bibitem{WangL_16a}
Limin Wang, Yuanjun Xiong, Zhe Wang, Yu Qiao, Dahua Lin, Xiaoou Tang, and Luc
  {Van Gool}.
\newblock Temporal segment networks: Towards good practices for deep action
  recognition.
\newblock In {\em ECCV}, 2016.

\bibitem{wang2017non}
Xiaolong Wang, Ross Girshick, Abhinav Gupta, and Kaiming He.
\newblock Non-local neural networks.
\newblock In {\em CVPR}, 2018.

\bibitem{winer2017automated}
David~R Winer and R~Michael Young.
\newblock Automated screenplay annotation for extracting storytelling
  knowledge.
\newblock In {\em Thirteenth Artificial Intelligence and Interactive Digital
  Entertainment Conference}, 2017.

\bibitem{wu2016google}
Yonghui Wu, Mike Schuster, Zhifeng Chen, Quoc~V Le, Mohammad Norouzi, Wolfgang
  Macherey, Maxim Krikun, Yuan Cao, Qin Gao, Klaus Macherey, et~al.
\newblock Google's neural machine translation system: Bridging the gap between
  human and machine translation.
\newblock {\em arXiv preprint arXiv:1609.08144}, 2016.

\bibitem{xie2018rethinking}
Saining Xie, Chen Sun, Jonathan Huang, Zhuowen Tu, and Kevin Murphy.
\newblock Rethinking spatiotemporal feature learning: Speed-accuracy trade-offs
  in video classification.
\newblock In {\em Proceedings of the European Conference on Computer Vision
  (ECCV)}, pages 305--321, 2018.

\bibitem{xu2019self}
Dejing Xu, Jun Xiao, Zhou Zhao, Jian Shao, Di Xie, and Yueting Zhuang.
\newblock Self-supervised spatiotemporal learning via video clip order
  prediction.
\newblock In {\em Proceedings of the IEEE Conference on Computer Vision and
  Pattern Recognition}, pages 10334--10343, 2019.

\bibitem{zhao2018sound}
Hang Zhao, Chuang Gan, Andrew Rouditchenko, Carl Vondrick, Josh McDermott, and
  Antonio Torralba.
\newblock The sound of pixels.
\newblock In {\em Proceedings of the European Conference on Computer Vision
  (ECCV)}, pages 570--586, 2018.

\bibitem{zhao2017slac}
Hang Zhao, Zhicheng Yan, Heng Wang, Lorenzo Torresani, and Antonio Torralba.
\newblock {SLAC}: A sparsely labeled dataset for action classification and
  localization.
\newblock {\em arXiv preprint arXiv:1712.09374}, 2017.

\bibitem{zhou2018towards}
Luowei Zhou, Chenliang Xu, and Jason~J Corso.
\newblock Towards automatic learning of procedures from web instructional
  videos.
\newblock In {\em Thirty-Second AAAI Conference on Artificial Intelligence},
  2018.

\bibitem{zhu2015aligning}
Yukun Zhu, Ryan Kiros, Rich Zemel, Ruslan Salakhutdinov, Raquel Urtasun,
  Antonio Torralba, and Sanja Fidler.
\newblock Aligning books and movies: Towards story-like visual explanations by
  watching movies and reading books.
\newblock In {\em Proceedings of the IEEE international conference on computer
  vision}, pages 19--27, 2015.

\end{thebibliography}

}
\clearpage
\appendix
We include additional details and results for training the Speech2Action model in Sec. \ref{supp:s2a}. In Sec. \ref{supp:mining}, we show more results for the techniques used to mine training samples -- i.e.\ the Keyword Spotting Baseline and the \texttt{Speech2Action} model. Finally, we show results on the UCF101~\cite{soomro2012ucf101} dataset in Sec. \ref{supp:ucf}.
\section{\texttt{Speech2Action} model}\label{supp:s2a}
\subsection{Screenplay Parsing} \label{supp:scripts}

We follow the grammar created by Winer et al.~\cite{winer2017automated} which is based on `The Hollywood Standard'~\cite{riley2009hollywood}, an authoritative guide to screenplay writing, to parse the screenplays and separate out various script elements. The tool uses spacing, indentation, capitalization and punctuation to parse screenplays into the following four different elements: \\
1.\ Shot Headings -- These are present at the start of each scene or shot, and may
give general information about a scene’s location, type of
shot, subject of shot, or time of day, e.g.\  {\tt INT. CENTRAL PARK - DAY} \\
2.\ Stage Direction -- This is the stage direction that is to be given to the actors. This contains the \textit{action} information that we are interested in, and is typically a paragraph containing many sentences,
e.g.\  {\tt Nason and his guys fight the fire. They
are CHOKING on smoke. PAN TO Ensign
Menendez, leading in a fresh contingent
of men to join the fight. One of them
is TITO.} \\
3.\ Dialogue -- speech uttered by each character, e.g.\  {\tt INDY: Get down!}  \\
4.\ Transitions -- may appear at the end of a scene, and indicate how
one scene links to the next, e.g.\  {\tt HARD CUT TO:} \\

\begin{figure*}[h]
\begin{center}
\includegraphics[width=1\textwidth]{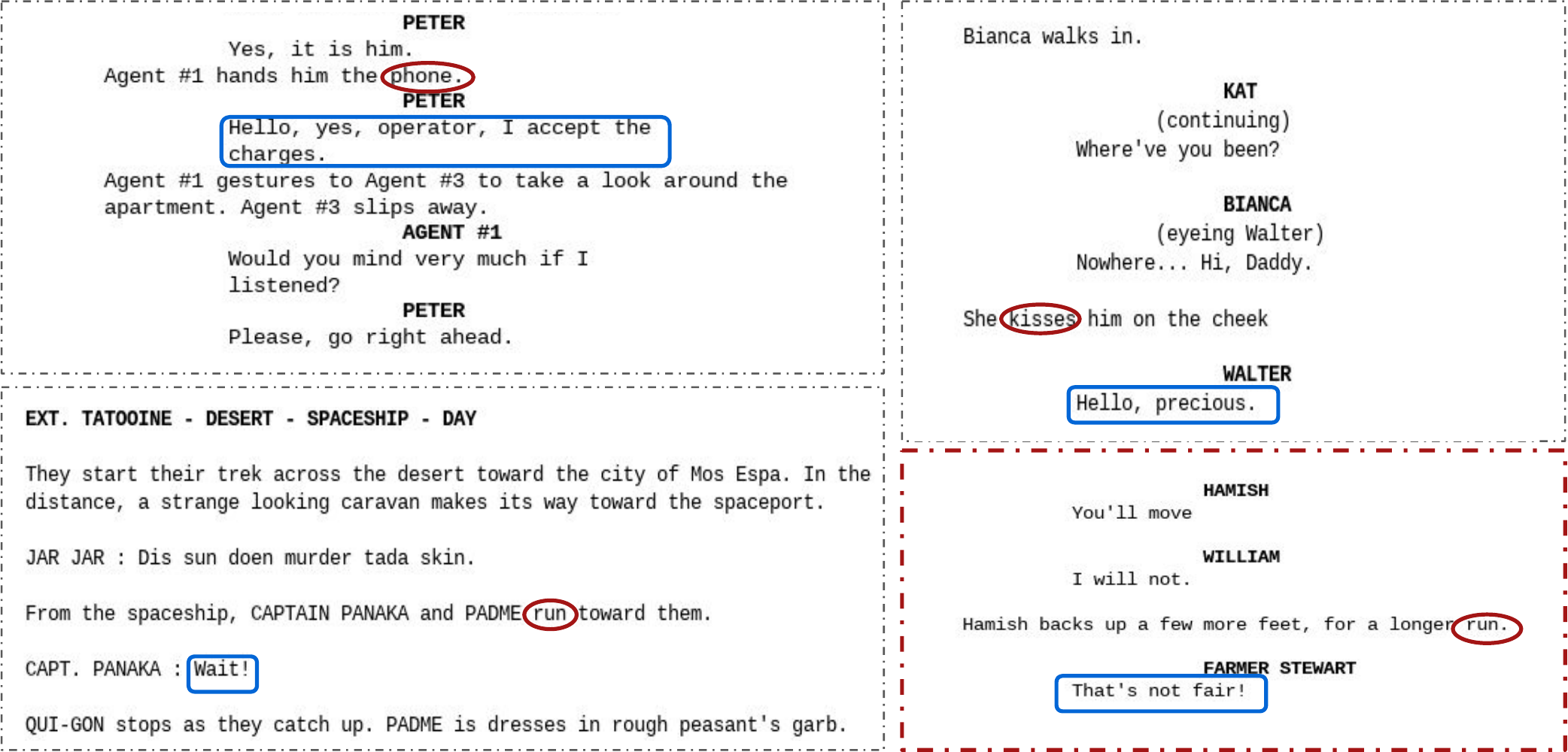}
\end{center}
   \caption{Examples of speech and verb action pairs obtain from screenplays. In the bottom row (right) we show a possibly negative speech and verb pair, i.e.\ the speech segment \textit{That's not fair!} is assigned the action verb `run', whereas it is not clear that these two are correlated. }
\label{fig:screenplays}
\end{figure*}

In this work we only extract 2.\ Stage Direction, and 3.\ Dialogue. After mining for verbs in the stage directions, we then search for the nearest section of dialogue (either before or after) and assign each sentence in the dialogue with the verb class label (see Fig.\ \ref{fig:screenplays} for examples of verb-speech pairs obtained from screenplays).

\begin{figure}
\begin{center}
\includegraphics[width=0.5\textwidth]{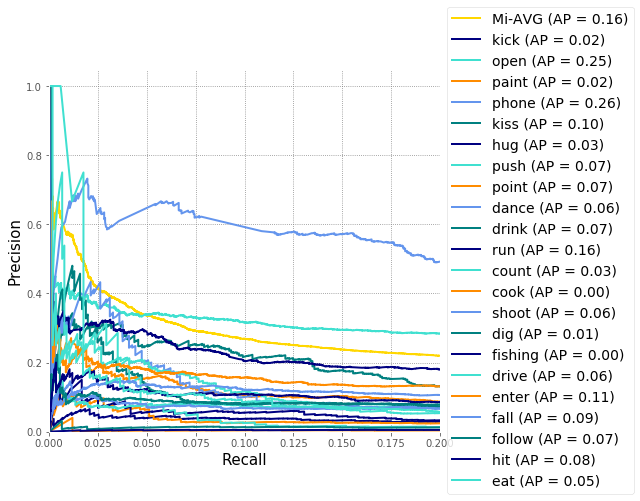}
\end{center}
   \caption{PR curves on the validation set of the IMSDb dataset for the \texttt{Speech2Action} model. Since the validation set is noisy, we are only interested in performance in the low recall, high precision setting. Note how some classes -- `phone', `open' and `run' perform much better than others.}
\label{fig:PR}
\end{figure}
\begin{table*}[h]
\centering 
\begin{footnotesize}
\begin{tabular}{c l c c l } 
\toprule
 \cellcolor{salmon}&   \cellcolor{salmon}why didn't you return my phone calls? && \cellcolor{mistyrose}&    \cellcolor{mistyrose}they were both undone by true love's kiss.         \\ 
 \cellcolor{salmon}&   \cellcolor{salmon}you each get one phone call && \cellcolor{mistyrose}&  \cellcolor{mistyrose}good girls don't kiss and tell.            \\
\cellcolor{salmon}phone &   \cellcolor{salmon}i already got your phone line set up. &&  \cellcolor{mistyrose} kiss &    \cellcolor{mistyrose}kiss my a**         \\
\cellcolor{salmon} &   \cellcolor{salmon}	but my phone died, so just leave a message, okay? && \cellcolor{mistyrose}&     \cellcolor{mistyrose}it was our first kiss.         \\
 \cellcolor{salmon}&   \cellcolor{salmon}i'm on the phone.. && \cellcolor{mistyrose}&  \cellcolor{mistyrose}i mean, when they say, "i'll call you," that's the kiss of death.              \\
 \cellcolor{salmon} &   \cellcolor{salmon}we're collecting cell phones, surveillance tapes, video we can find. && \cellcolor{mistyrose}&    \cellcolor{mistyrose}i had to kiss jace.              \\

\cellcolor{lavenderblue} &she went to the dance with Harry Land \cellcolor{lavenderblue} && \cellcolor{azure}&     \cellcolor{azure}against a top notch britisher, you'll be eaten alive.          \\ 
\cellcolor{lavenderblue} &\cellcolor{lavenderblue}do you wanna dance? && \cellcolor{azure}&    \cellcolor{azure}eat my dust, boys!            \\
\cellcolor{lavenderblue}dance &   \cellcolor{lavenderblue}and the dance of the seven veils? && \cellcolor{azure} eat & \cellcolor{azure}ate something earlier.       \\
\cellcolor{lavenderblue} &\cellcolor{lavenderblue}what if i pay for a dance? &&\cellcolor{azure}&    \cellcolor{azure}i can't eat, i can't sleep.      \\
\cellcolor{lavenderblue}&\cellcolor{lavenderblue}the dance starts in an hour. &&\cellcolor{azure}& \cellcolor{azure}you must eat the sardines tomorrow.             \\
 \cellcolor{lavenderblue} & \cellcolor{lavenderblue}just dance. &&\cellcolor{azure}&\cellcolor{azure}i ate bad sushi.              \\
\cellcolor{lemonchiffon} &\cellcolor{lemonchiffon}are you drunk? && \cellcolor{pistachio} &     \cellcolor{pistachio}and you can add someone to an email chain at any point.          \\ 
\cellcolor{lemonchiffon} &\cellcolor{lemonchiffon}my dad would be drinking somewhere else. &&\cellcolor{pistachio}&\cellcolor{pistachio}she's got a point, buddy.            \\
\cellcolor{lemonchiffon}drink &\cellcolor{lemonchiffon}you didn't drink the mold. && \cellcolor{pistachio} point &\cellcolor{pistachio}the point is, they're all having a great time.         \\
\cellcolor{lemonchiffon} &\cellcolor{lemonchiffon}let's go out and drink. &&\cellcolor{pistachio}&\cellcolor{pistachio}didn't advance very far, i think, is mark's point.         \\
\cellcolor{lemonchiffon} &\cellcolor{lemonchiffon}super bowl is the super bowl of drinking. &&\cellcolor{pistachio}& \cellcolor{pistachio}you made your point.            \\
\cellcolor{lemonchiffon}  &\cellcolor{lemonchiffon}i don't drink, i watch my diet, but no. &&\cellcolor{pistachio}&\cellcolor{pistachio} beside the point!                \\
\bottomrule
\end{tabular}
\end{footnotesize}
\caption{\textbf{Examples of speech samples for six verb categories labelled with the keyword spotting baseline.} Each block shows the action 
verb on the left, and the speech samples on the right.  Since we do not need to use the movie screenplays for this baseline, unlike \texttt{Speech2Action} (results in Table.\ 2 of the main paper), we show examples of transcribed speech obtained directly from the unlabelled corpus. Note how the speech labelled with the verb `point' is indicative of a different semantic meaning to the physical action of `pointing'. }
\label{table:baselineexamples} 
\end{table*}
\subsection{PR Curves on the Validation Set of the IMSDb Data}\label{supp:PR}
We show precision-recall curves on the val set of the IMSDb dataset in Fig.~\ref{fig:PR}. Note how classes such as `run' and `phone' have a much higher recall for the same level of precision.  

We select thresholds for the \texttt{Speech2Action} model using a greedy search as follows:
(1) We allocate the retrieved samples into discrete precision buckets (30\%-40\%, 40\%-50\%, etc.), using thresholds obtained from the PR curve mentioned above;
(2) For different actions, we adjust the buckets to make sure the number of training examples are roughly balanced for all classes;
(3) For classes with low precision, in order to avoid picking uncertain and hence noiser predictions, we only select examples that had a precision above 30\%+.

\begin{figure*}[ht]
  \centering
  \begin{minipage}[b]{0.45\textwidth}
    \includegraphics[width=\textwidth]{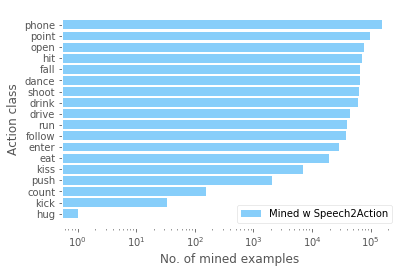}
    \caption{\textbf{Distribution of training clips mined using
   \texttt{Speech2Action}}. We show the distribution for \textit{all} 18 verb classes. It is difficult to mine clips for the actions `hug' and `kick', as these are often confused with `kiss' and `hit'. }
   \label{fig:trainingdists2a}
  \end{minipage}
  \hfill
  \begin{minipage}[b]{0.45\textwidth}
    \includegraphics[width=\textwidth]{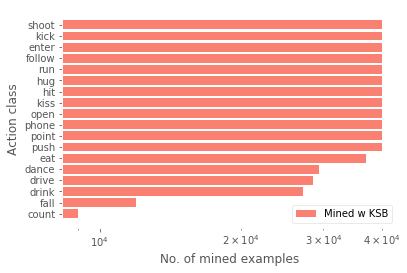}
    \caption{\textbf{Distribution of training clips mined using
   the Keyword Spotting baseline}. We show the distribution for \textit{all} 18 verb classes. We cut off sampling at 40K samples for twelve classes in order to prevent too much of a class imbalance.}
   \label{fig:trainingdistksb}
  \end{minipage}
\end{figure*}

The number of retrieved samples per class can be seen in Fig.~\ref{fig:trainingdists2a}. The number of retrieved samples for `phone' and `open' at a precision value of 30\% are in the millions (2,272,906 and 31,657,295 respectively), which is why we manually increase the threshold in order to prevent a large class-imbalance during training. We reiterate here once again that this evaluation is performed purely on the basis of the proximity of speech to verb class in the stage direction of the movie screenplay (Fig.\ \ref{fig:screenplays}), and hence it is \textit{not} a perfect ground truth indication of whether an action will actually be performed in a \textit{video} (which is impossible to say only from the movie scripts). We use the stage directions in this case as \textit{pseudo} ground truth. There are many cases in the movie screenplays where verb and speech pairs could be completely uncorrelated (see Fig.\ \ref{fig:screenplays}, bottom--right for an example.)

\begin{figure*}[h]
\begin{center}
\includegraphics[width=1\textwidth]{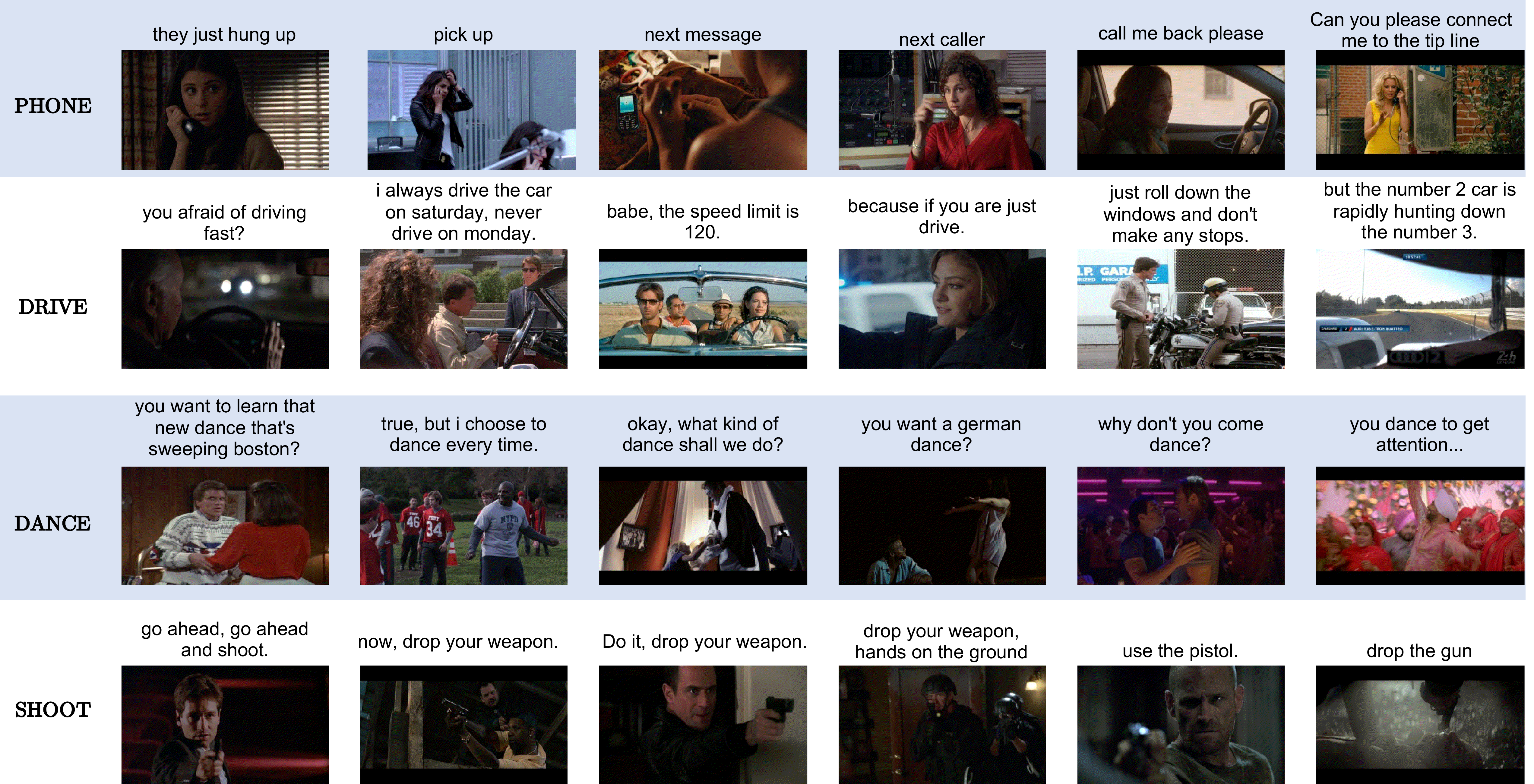}
\end{center}
   \caption{\textbf{Examples of clips mined automatically using the \texttt{Speech2Action} model applied to \textit{speech alone} for 4 AVA classes.} We show only a single frame from each video. Note the diversity in \textit{object} for the category `[answer] phone' (first row, from left to right) a landline, a cell phone, a text message on a cell phone, a radio headset, a carphone, and a payphone, in \textit{viewpoint} for the category `drive' (second row) including behind the wheel, from the passenger seat, and from outside the car, and in \textit{background} for the category `dance' (third row, from left to right) inside a home, on a football pitch, in a tent, outdoors, in a club/party and at an Indian wedding/party. }
\label{fig:mining_more}
\end{figure*}

\section{Mining Techniques}\label{supp:mining}

\subsection{Keyword Spotting Baseline}
In this section we provide more details about the Keyword Spotting Baseline (described in Sec. 4.2.2  of the main paper). The total number of clips mined using the Keyword Spotting Baseline is 679,049. We mine all the instances of speech containing the verb class, and if there are more than 40K samples, we randomly sample 40K clips. The reason we cap samples at 40K is to prevent overly unbalanced classes.  Examples of speech labelled with this baseline for 6 verb classes can be seen in Table \ref{table:baselineexamples}. There are two ways in which our learned \texttt{Speech2Action} model is theoretically superior to this approach: \\
(1) Many times the speech correlated with a particular action does not actually contain the action verb itself e.g.\ \textit{`Look over there'} for the class `point'. \\
(2) There is no \textit{word-sense disambiguation} in the way the speech segments are mined, i.e. \textit{`Look at where I am pointing'} vs \textit{`You've missed the point'}. Word-sense disambiguation is the task of identifying which sense of a word is used in a sentence when a word has multiple meanings. This task tends to be more difficult with verbs than nouns because verbs have more senses on average than nouns and may be part of a multiword phrase~\cite{corro2014werdy}.

\subsection{Mined Examples}
The distribution of mined examples per class for all 18 classes, using the \texttt{Speech2Action} model and the Keyword Spotting baseline can be seen in Figures \ref{fig:trainingdists2a} and \ref{fig:trainingdistksb}. We note that it is very difficult to mine examples for actions `hug' and `kick', as these are often accompanied with speech similar to that accompanying `kiss' and `hit'.

We show more examples of automatically mined video clips from unlabelled movies using the \texttt{Speech2Action} model in Fig.~\ref{fig:mining_more}. Here we highlight in particular the diversity of video clips that are mined using simply speech \textit{alone}, including diversity in objects, viewpoints and background scenes.

\section{Results on UCF101}\label{supp:ucf}
In this section we show the results of pretraining on our mined video examples and then finetuning on the UCF101 dataset~\cite{soomro2012ucf101}, following the exact same procedure described in Sec. 5.1 of the main paper. UCF101~\cite{soomro2012ucf101} is a dataset of 13K videos downloaded from YouTube spanning over 101 human action classes. Our results follow a similar trend to those on HMDB51, pretraining on samples mined using \texttt{Speech2Action} (81.4\%) outperforms training from scratch (74.2\%) and pretraining on samples obtained using the keyword spotting basline (77.4\%). We note here, however, that it is much harder to tease out the difference between various styles of pretraining on this dataset, because it is more saturated than HMDB51 (training from scratch already yields a high accuracy of 74.2\%, and pretraining on Kinetics largely solves the task, with an accuracy of 95.7\%).  

\begin{table}[ht]
\centering 
\footnotesize{
\begin{tabular}{l l l  l  r}
\toprule
Method & Architecture  & Pre-training & Acc. \\
% Shuffle\&Learn (Misra {\it et al.}~\cite{misra2016shuffle}) & AlexNet~\cite{krizhevsky2012imagenet} & Scratch & 13.3 \\
% Shuffle\&Learn (Misra {\it et al.}~\cite{misra2016shuffle}) & AlexNet~\cite{krizhevsky2012imagenet} & Tuple verify~\cite{misra2016shuffle} &  18.1 \\
% Shuffle\&Learn (Misra {\it et al.}~\cite{misra2016shuffle}) & AlexNet~\cite{krizhevsky2012imagenet}  & ImageNet &  28.5 \\
% Two-stream (RGB)~\cite{simonyan2014two} & VGG-M~\cite{Simonyan_14a} & 1 & ImageNet & - & 40.5 \\
% C3D~\cite{carreira2017quo} & Custom  & Scratch & 24.3 & - \\
% \midrule
% LSTM~\cite{carreira2017quo} & BN-Inception~\cite{ioffe2015batch} & - & ImageNet & 36.0 & - \\
% Two stream (RGB)~\cite{carreira2017quo} & BN-Inception~\cite{ioffe2015batch} & 1 & ImageNet & 43.2 & - \\
% I3D (RGB)~\cite{carreira2017quo} & BN-Inception~\cite{ioffe2015batch} & 64 & ImageNet & 49.8 & - \\
% \midrule

\midrule
Shuffle\&Learn~\cite{misra2016shuffle}$\star$ & S3D-G (RGB) & UCF101$\dagger$~\cite{soomro2012ucf101}& 50.2\\
OPN \cite{lee2017unsupervised}  & VGG-M-2048  & UCF101$\dagger$~\cite{soomro2012ucf101}  & 59.6 \\
ClipOrder \cite{xu2019self} & R(2+1)D  & UCF101$\dagger$~\cite{soomro2012ucf101}   & 72.4 \\
Wang et al. \cite{wang2019self} & C3D  & Kinetics$\dagger$~\cite{soomro2012ucf101}   & 61.2 \\
3DRotNet \cite{jing2018self}$\star$ & S3D-G (RGB)  & Kinetics$\dagger$   & 75.3  \\
DPC \cite{han2019video} & 3DResNet18 & Kinetics$\dagger$ & 75.7 \\
CBT \cite{sun2019contrastive} & S3D-G (RGB) & Kinetics$\dagger$   & 79.5 \\
\midrule 
DisInit (RGB) \cite{girdhar2019distinit} & R(2+1)D-18 \cite{tran2018closer} & Kinetics$**$  & 85.7 \\  
Korbar et al \cite{korbar2018cooperative} & I3D (RGB) & Kinetics$\dagger$  & 83.7 \\  
% Ours (RGB) & S3D-G (RGB) & 64 & \texttt{KSB-mined} & \textbf{**} & -- \\
\midrule 
- & S3D-G (RGB) & Scratch & 74.2 \\ 
Ours & S3D-G (RGB)  & \texttt{KSB-mined}  & 77.4 \\
Ours & S3D-G (RGB)  & \texttt{S2A-mined}  & 81.4 \\
\midrule
Supervised pretraining & S3D-G (RGB)  & ImageNet & 84.4  \\
Supervised pretraining & S3D-G (RGB)  & Kinetics & 95.7  \\
\bottomrule
\end{tabular}}
\caption{\textbf{Comparison with previous pre-training strategies for action classification on UCF101.} Training on videos labelled with \texttt{Speech2Action} leads to a 7\% improvement over training from scratch and outperforms previous self-supervised works. It also performs competitively with other weakly supervised works.
\textbf{KSB-mined:} video clips mined using the keyword spotting baseline.
\textbf{S2A-mined:} video clips mined using the \texttt{Speech2Action} model. $\dagger$videos without labels. **videos with labels distilled from ImageNet.  When comparing to~\cite{korbar2018cooperative}, we report the number achieved by their I3D (RGB only) model which is the closest to our architecture. For $\star$, we report the reimplementations by \cite{sun2019contrastive} using the S3D-G model (same as ours). For the rest, we report performance directly from the original papers.}

% Note that we do not compare to Kinetics pre-trained models.
% Using Kinetics for pre-training with I3D~\cite{carreira2017quo} gets 74.3\% and 95.1\% 3-split avg on HMDB and UCF, but is not comparable to our unsupervised approach which does not use those labels.
\label{tab:sota-ucf}
\end{table} 

\end{document}